\documentclass[conference]{IEEEtran}

\input{xhc.sty}

\usepackage{lipsum}
\usepackage{multirow}

\begin{document}

\title{Learning Dynamical Demand Response Model in Real-Time Pricing Program}

\author{\IEEEauthorblockN{Hanchen Xu\IEEEauthorrefmark{1}\IEEEauthorrefmark{2},
Hongbo Sun\IEEEauthorrefmark{2},
Daniel Nikovski\IEEEauthorrefmark{2}, 
Kitamura Shoichi \IEEEauthorrefmark{3}, and
Kazuyuki Mori \IEEEauthorrefmark{3}}
\IEEEauthorblockA{\IEEEauthorrefmark{1}Department of Electrical and Computer Engineering, University of Illinois,
Urbana, IL 61801, USA}
\IEEEauthorblockA{\IEEEauthorrefmark{2}Mitsubishi Electric Research Laboratories, 201 Broadway, Cambridge, MA 02139, USA}
\IEEEauthorblockA{\IEEEauthorrefmark{3}Advanced Technology R\&D Center, Mitsubishi Electric Corporation, Hyogo 661-8661, Japan}
Email: hxu45@illinois.edu, hongbosun@merl.com, nikovski@merl.com
}

\maketitle

\begin{abstract}
Price responsiveness is a major feature of end use customers (EUCs) that participate in demand response (DR) programs, and has been conventionally modeled with static demand functions, which take the electricity price as the input and the aggregate energy consumption as the output.
This, however, neglects the inherent temporal correlation of the EUC behaviors, and may result in large errors when predicting the actual responses of EUCs in real-time pricing (RTP) programs.
In this paper, we propose a dynamical DR model so as to capture the temporal behavior of the EUCs.
The states in the proposed dynamical DR model can be explicitly chosen, in which case the model can be represented by a linear function or a multi-layer feedforward neural network, or implicitly chosen, in which case the model can be represented by a recurrent neural network or a long short-term memory unit network.
In both cases, the dynamical DR model can be learned from historical price and energy consumption data.
Numerical simulation illustrated how the states are chosen and also showed the proposed dynamical DR model significantly outperforms the static ones.
\end{abstract}

\section{Introduction}

Demand response (DR) has proven to be a successful paradigm that emerges in the context of smart grid \cite{wang2017values, rahimi2010demand, medina2010demand}.
The DR enables the participation of demand-side  resources, including commercial buildings \cite{zhang2016conceptual, zhang2017self}, industrial loads \cite{zhang2018demand}, electric vehicles \cite{zhang2018dynamic}, in maintaining the supply-demand-balance in power systems through some economic means such as incentives or price signals \cite{wang2017values}.
A variety of DR programs have been developed so far on both the wholesale market level and the retail market level.
On the wholesale level, qualified demand-side resources can offer in almost all markets including capacity market, energy market, as well as ancillary markets, and will be cleared by the independent system operator (ISO) in the same way as the conventional generating resources.
DR programs on the retail level, however, have more variety than those on the wholesale level.
Yet, the retail DR programs can, by-and-large, be categorized into two classes---incentive based and price based.
The former includes, e.g, direct load control and some interruptible programs, while the latter includes the real-time pricing (RTP), time-of-use-pricing, critical peaking pricing.

In price-based DR programs, the load serving entity (LSE)---the operator of the retail market---needs to design effective pricing schemes such that certain objective such as minimizing load fluctuation \cite{sun2016mitigating}, maximizing profit \cite{xu2018optimal} or utility \cite{kim2016dynamic}, can be achieved.
Particularly in RTP, the LSE sends out a price signal for every time interval, and the end use customers (EUCs) then respond to the price signal by adjusting their energy consumptions.
It is therefore critical for the LSE to know the price response characteristics of the EUCs in order to make informed decisions on the price signal sent to the EUCs.
Conventionally, price response characteristics are modeled using a static function such as linear, exponential, logarithmic, and potential functions \cite{yousefi2011optimal}.
These static functions, however, neglects the inherent temporal correlation of the EUC behaviors, and as we will see in the later part of the paper, may result in large errors when predicting the actual responses of EUCs in RTP programs.

To explicitly capture the temporal behavior of the EUCs, we propose a dynamical DR model which has states that evolve over time.
The states in the proposed dynamical DR model keep necessary information from previous time intervals, and allows more accurate prediction of the EUCs' response.
These states can be explicitly chosen, in which case the model can be represented by a linear function or a multi-layer feedforward neural network (FNN), or implicitly chosen, in which case the model can be represented by a recurrent neural network (RNN) or a long short-term memory (LSTM) unit network.
The proposed dynamical DR model can be learned from historical price and energy consumption data.
In particular, the states can be determined empirically from the historical data.
We will also show through numerical simulation that the proposed dynamical DR model significantly outperforms the conventional static models.

The remaining part of this paper is organized as follows. Section II reviews decision process of EUCs and proposes a dynamical DR model that is based on available data. The neural network representations of the dynamical DR model are developed in Section III, and results from numerical simulation are presented in Section IV. Some concluding remarks are made in Section V.

\section{Demand Response Model} \label{sec:model}

In this section, we first briefly review how a myopic EUC may determine its energy consumption at a given electricity price.
In light of the deficiencies in static DR models, we propose a dynamical DR model in the end.
Throughout this paper, we use a subscript $t$ to denote the time interval, and assume one day is decomposed into $T$ segments.

\subsection{Decision Process of End Use Customers}

In a RTP program, the LSE sends out a energy price for time interval $t$, denoted by $p_t$, before the beginning of that time interval.
Each EUC then responds to the price signal by optimizing its energy consumption in a way that maximizes its overall benefit or equivalently, minimizes its overall cost.
While there exist many detailed models for EUCs', a generic model that is agnostic to the underlying components is to model the \textit{energy demand} and \textit{energy consumption} separately, where the energy demand is the needed energy while the energy consumption is the actual consumed energy \cite{conejo2010real, kim2016dynamic}.

Let $\calK = \{1, \cdots, K\}$ denote the set of EUCs served by the LSE.
Let $e_t^{k, d}$ and $e_t^{k, c}$ denote the energy demand and energy consumption of EUC $k \in \calK$ at time interval $t$, respectively.
A myopic EUC finds its optimal energy consumption by solving the following utility maximization problem \cite{kim2016dynamic}:
\begin{subequations} \label{eq:euc}
	\begin{align} \label{eq:euc_obj}
	\maximize_{e_t^{k, c}}~\beta_t^k(e_t^{k, d}, e_t^{k, c}) - p_t e_t^{k, c},
	\end{align}
	\text{subject to}
	\begin{align} \label{eq:euc_dynamics}
	e_{t+1}^{k, d} = \alpha_t^k (e_t^{k, d} - e_t^{k, c}) + \xi_{t+1}^k,
	\end{align}
	\begin{align} \label{eq:euc_feasible}
	e_{t}^{k, c} \in \calE_{t}^{k, c},
	\end{align}
\end{subequations}
where $\beta_t^k(\cdot)$ is the benefit function, which gives the benefit of the EUC at certain energy demand and energy consumption, $\alpha_t^k$ is the backlog rate that represents the percentage of unmet energy demand that is carried over to the next time interval, $\xi_{t+1}^k$ is a random variable that models the new demand, $\calE_{t}^{k, c}$ is the feasible set of energy consumption.

From the perspective of the LSE, the collective behavior of all EUCs are of more interest since its total profit demands on the aggregate energy consumption.
The DR model that is of interest to the LSE is essentially a mapping from the price to the aggregate energy consumption as follow:
\begin{align} \label{eq:mapping}
	p_t \longmapsto e_t^c := \sum_{k \in \calK} e_t^{k, c}.
\end{align}

\subsection{Dynamical Demand Response Model}

The mapping from price to energy consumption is conventionally modeled using a static function such as linear, exponential, logarithmic, and potential functions \cite{yousefi2011optimal}, which can be conceptually expressed as:
\begin{align}
e_t^c = \psi(p_t).
\end{align}
These static functions generally work well on wholesale level since the price response characteristics of DR resources in the wholesale market are mainly determined by their offers.
Once the energy demand are cleared for each time interval, the DR resources will be controlled such that their energy consumptions follow the schedule, since otherwise they will incur costs.
Therefore, given a certain price, the energy consumption can be determined from the relative static offer stacks directly, i.e., the mapping is relatively static.

However, in a RTP problem, the EUCs do not have the responsibility to follow any energy consumption profile.
Indeed, it is obvious from \eqref{eq:euc} that given a price, the energy consumption of an EUC will depend on its current energy demand, which itself depends on the previous energy consumption.
Therefore, the mapping in \eqref{eq:mapping} is dynamical.

In light of this observation, we propose a dynamical model for the DR model for the EUCs in a RTP program as follows:
\begin{align} \label{eq:dr_model}
e_t^c = \varphi(\bm{s}_t, p_t),
\end{align}
where $\bm{s}_t$ is a state vector that captures all the factors besides the price that impact the aggregate energy consumption, and evolves over time given $p_t$.

While it is difficult to identify the ``perfect" state vector, we may still be able to construct one that is good enough in the sense it gives good prediction accuracy of the aggregate demand.
From a practical view, the state vector can be compromised of elements from the set of information that is available to the LSE at time interval $t$, denoted by $\calI_t = \{(p_\tau, e_\tau^c)~\forall \tau<t, p_t, t\}$.
There are two potential approaches to construct such a state vector, a \textit{direct} approach, and an \textit{indirect} approach.
In the direct approach, we select the state vector to be
\begin{align} \label{eq:state1}
\bm{s}_t = (p_{t-n}, e_{t-n}^c, \cdots, p_{t-1}, e_{t-1}^c, t\bmod T),
\end{align}
where $n$---referred to as the \textit{order} of the DR model---is a parameter that can be determined from the historical data.
We adopt the convention that $\bm{s}_t = \{t\bmod T\}$ when $n=0$.
In the indirect approach, the state vector in \eqref{eq:state1} is further transformed in a series of nonlinear operations.
With the state vector and the price information, we can predict the aggregate energy consumption using supervised learning techniques.

\section{Neural Network Representations of Dynamical Demand Response Model} \label{sec:nn}

In this section, we develop neural network representations for the dynamical DR model.
A multi-layer FNN is applied in the direct approach, and a multi-layer RNN is applied in the indirect approach.

\begin{figure}[!t]
	\centering
	\includegraphics[scale=0.6]{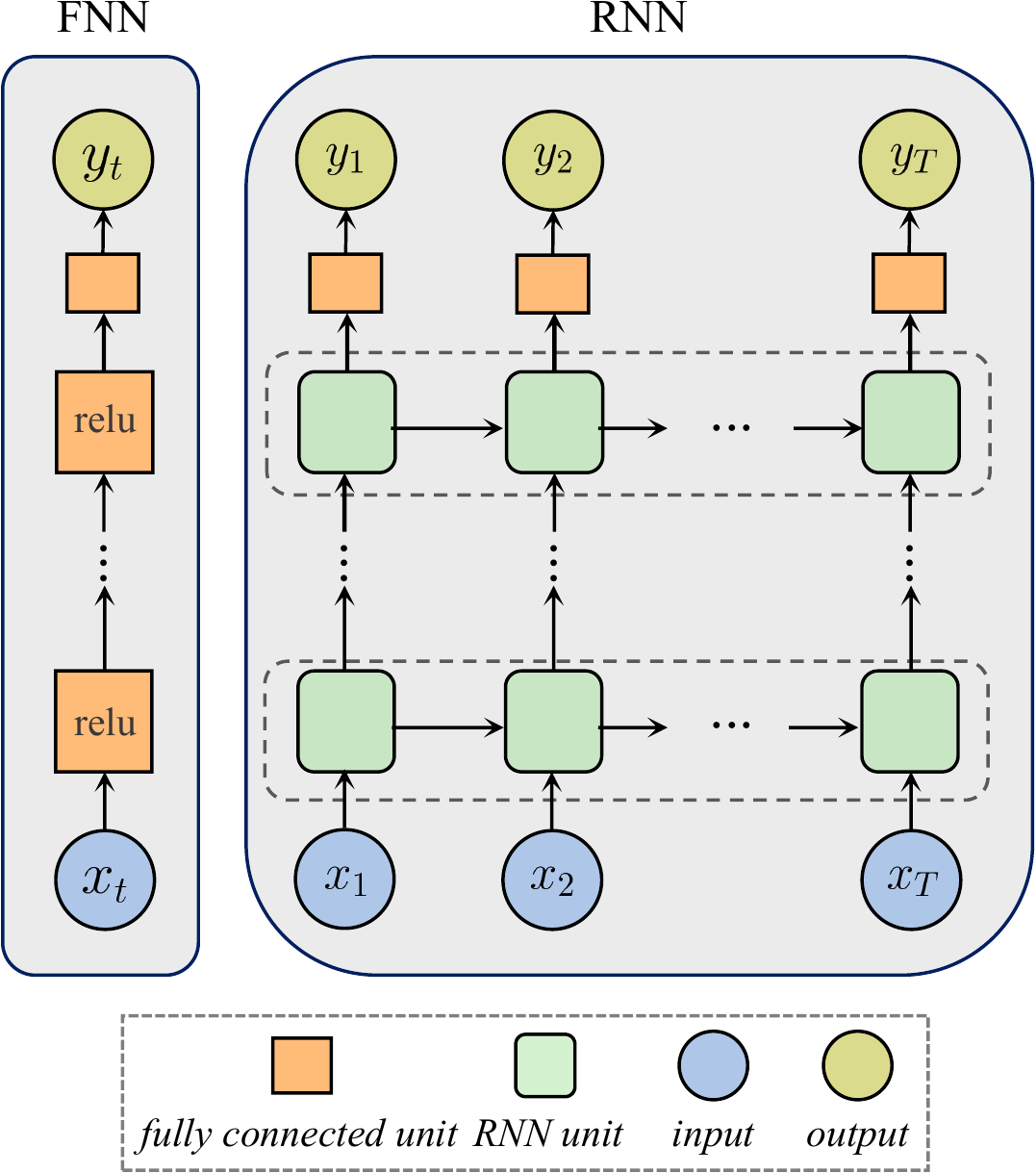}
	\caption{Neural networks for dynamical demand response model.}
	\label{fig:rnn}
\end{figure}

\subsection{Direct Approach Using Feedforward Neural Network}

When using the direct approach, the state vector is manually selected from the set of available information.
Therefore, the input vector and the output of the DR model are both readily available.
We can thus fit the dynamical DR model in \eqref{eq:dr_model} using, for example, a linear function as follows:
\begin{align}
e_t^c = \bm{w}^\top [\bm{s}_t^\top, p_t] + b,
\end{align}
where $\bm{w}$ is a weight vector, and $b$ is the bias, the superscript $\top$ denotes the transpose of a vector or matrix.

Also, we may represent $\varphi$ with a nonlinear function such as a multi-layer FNN, which consists of one input layer, $L$ hidden layer, and one output layer as illustrated in the left part in Fig. \ref{fig:rnn}.
The hidden layer $l$ takes an input vector $\bm{x}_t^{[l]}$, and computes a (hidden) output vector $\bm{h}_t^{[l]}$ according to
\begin{align}
\bm{h}_t^{[l]} = \mathrm{relu}(\bm{W}^{[l]} \bm{x}_t^{[l]} + \bm{b}^{[l]}),
\end{align}
where $\mathrm{relu}(\cdot)$ denotes a rectified linear unit function that is applied element-wise, $\bm{W}^{[l]}$ is a weight matrix, and $\bm{b}^{[l]}$ is a bias vector.
Note that the output vector of one hidden layer is the input vector for the next hidden layer, i.e, $\bm{x}^{[l+1]} = \bm{h}^{[l]}$, except the last hidden layer, the output of which is mapped to the output through a fully connected unit as follows:
\begin{align}
y_t = \bm{W} \bm{h}^{[L]} + \bm{b},
\end{align}
where $\bm{W}$ is a weight matrix, and $\bm{b}$ is a bias vector.
Note that $\bm{x}_t^{[1]} = (\bm{s}_t, p_t)$.
The multi-layer FNN can be trained using back-propagation such that the mean squared error between the predicted output $y_t$ and the true value $e_t^c$ is minimized, i.e., by minimizing the following loss function:
\begin{align} \label{eq:loss}
\ell = \frac{1}{m} \sum_{l=1}^m (y_t - e_t^c)^2.
\end{align}

\subsection{Indirect Approach Using Recurrent Neural Network}

Alternatively, the states can be implicitly constructed within the neural network, which leads to RNNs \cite{goodfellow2016deep}.
The right part in Fig. \ref{fig:rnn} illustrate a multi-layer RNN with one input layer, $L$ hidden layers, and one output layer.
The hidden state of the RNN unit in layer $l$, denoted by $\bm{h}_t^{[l]}$, is the input for RNN unit in the next layer as well as the input for itself at the next time step, as indicated by the arrows in Fig.\ref{fig:rnn}.
This RNN takes a sequence $\{\bm{x}_1, \cdots, \bm{x}_T\}$ as the input, and outputs a sequence $\{y_1, \cdots, y_T\}$.
Meanwhile, $L$ sequences of hidden states $\{\bm{h}_t^{[l]}\}$, $l=1, \cdots, L$, are generated along the trajectory, based on the following equations:
\begin{align}
\bm{h}_t^{[l]} &= \tanh(\bm{W}_{h}^{[l]} \bm{h}_{t-1}^{[l]} + \bm{W}_{x}^{[l]} \bm{x}_t^{[l]} + \bm{b}^{[l]}),
\end{align}
where $\tanh(\cdot)$ is applied element-wise, $\bm{W}_{h}$ and $\bm{W}_{x}$ are weight matrices, $\bm{b}$ is a bias vector.
Note that $\bm{h}_{-1}^{[l]}$ are initialized to zeros, $\bm{x}_t^{[l]} = \bm{h}_t^{[l-1]}$ for $l=2,\cdots, L$, and $\bm{x}_t^{[1]} = (\bm{s}_t, p_t)$.
The hidden states in RNN are dynamical since their values also depend on their previous values, while those in the FNN are static since their values purely depend on the inputs.
The output of the last hidden state vector is mapped to the output through a fully connected unit as in the case of multi-layer FNN.
The RNN can be trained by minimizing the same loss function as in \eqref{eq:loss} using backpropagation through time technique (see, e.g., \cite{werbos1990backpropagation}).
The input vector only has to include the most-recent information, i.e., when RNN is used, $n=1$ and $\bm{s}_t = \{p_{t-1}, e_{t-1}^c, t\bmod T\}$.

The key difference between the RNN and FNN in representing the dynamical DR model is that the FNN captures the temporal impacts by explicitly specifying a set of historical data as inputs, while the RNN keeps the temporal impacts by implicitly computing a dynamical hidden state.

\begin{figure}[!t]
	\centering
	\includegraphics[scale=0.6]{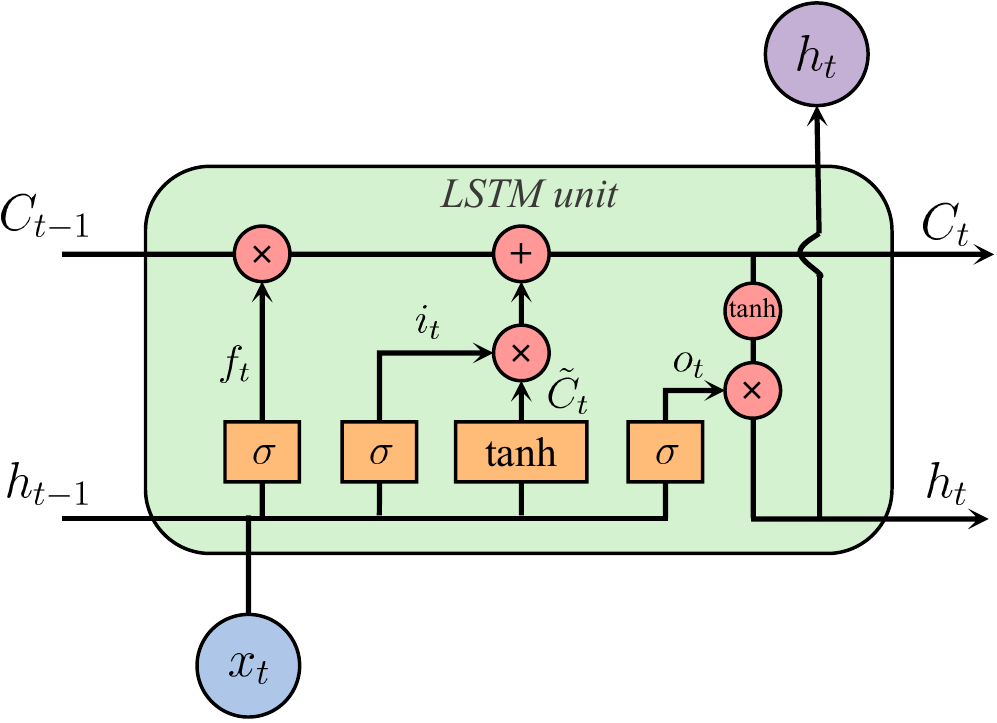}
	\caption{Structure of an LSTM unit \cite{understandLSTM}.}
	\label{fig:lstm}
\end{figure}

One of the deficiency of the basic RNN unit is the lack of ability to model long-term dependencies.
As a significant improvement over the basic RNN unit, the LSTM is proposed in \cite{hochreiter1997long}.
The structure of an LSTM unit is illustrated in Fig. \ref{fig:lstm}, in which $\sigma(\cdot)$ denotes a sigmoid function.
For the purpose of simplicity, we drop the superscript that indicates the layer and focus on structure inside one LSTM unit.
The LSTM unit introduces a new hidden state vector $\bm{C}_t$, which is used to keep long-term memories.
The LSTM unit works as follows.
First, a forget gate vector $\bm{f}_t$, an information gate vector $\bm{i}_t$, and an output gate vector $\bm{o}_t$ is computed from previous hidden state $\bm{h}_{t-1}$ and new input vector $\bm{x}_t$ as follows:
\begin{align}
\bm{f}_t &= \sigma(\bm{W}_{fh} \bm{h}_{t-1} + \bm{W}_{fx} \bm{x}_{t} + \bm{b}_f), \\
\bm{i}_t &= \sigma(\bm{W}_{ih} \bm{h}_{t-1} + \bm{W}_{ix} \bm{x}_{t} + \bm{b}_i), \\
\bm{o}_t &= \sigma(\bm{W}_{oh} \bm{h}_{t-1} + \bm{W}_{ox} \bm{x}_{t} + \bm{b}_o).
\end{align}
Then, the two hidden state vectors are updated as follows:
\begin{align}
\tdbdC_t &= \tanh(\bm{W}_{Ch} \bm{h}_{t-1} + \bm{W}_{Cx} \bm{x}_{t} + \bm{b}_C),\\
\bm{C}_t &= \bm{f}_t \circ \bm{C}_{t-1} + \bm{i}_t \circ \tdbdC_t,\\
\bm{h}_t &= \bm{o}_t \circ \tanh(\bm{C}_t),
\end{align}
where $\circ$ represents element-wise multiplication.
This structure has proven to be very effective in capturing long-term temporal dependencies, and therefore, is expected to outperform the basic RNN unit when representing the dynamical DR model.
The multi-layer LSTM network is similar to the RNN in Fig. \ref{fig:rnn}.
We simply replace the RNN unit with the LSTM unit.

\section{Simulation Results} \label{sec:simu}

In this section, we numerically investigate the performance of the proposed dynamical DR model under the direct and indirect approaches.

\subsection{Simulation Setup}

\begin{figure}[!t]
	\centering
	\includegraphics[scale=0.6]{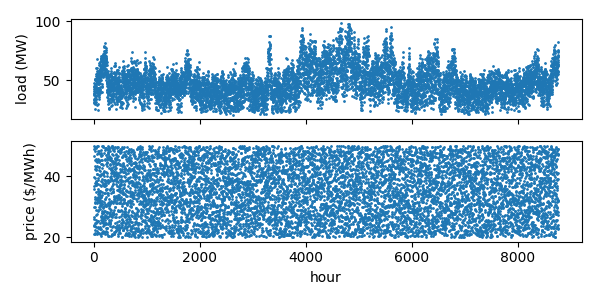}
	\caption{Simulated historical energy consumption and price data.}
	\label{fig:data}
\end{figure}

Assume $T=24$, i.e., each time interval covers one hour.
We assume the benefit function has a quadratic form, the parameter in which is a constant, i.e.,
\begin{align}
\beta_t^k(e_t^{k, d}, e_t^{k, c}) = \rho^k (e_t^{k, d} - e_t^{k, c})^2.
\end{align}
The feasible set is $\calE_{t}^{k, c} = \{e_t^{k, c} \geq 0.5  e_t^{k, d}\}$.
The backlog rate is uniformly sampled from $[0, 1]$.
The new demand is generated according the following procedures.
First, the peak demand of each EUC is sampled uniformly from $[0.1, 2]$ MW, then the new demand of the EUC is computed as the product of the peak demand, the normalized annual load profile from one of the zones in PJM in 2017 \cite{PJM}, and a Gaussian random variable with a mean of $1$ and a standard devision of $0.1$.
The value of $\rho^k$ is taken as the ratio of $-100$ \$/MWh$^2$ and the EUC's peak demand.
It is chosen in this way such that the responses from EUCs are reasonable.
In reality, the value of $\rho^k$ is a choice of the EUCs.
The number of EUCs served by the LSE is $100$.
The historical data are simulated using the parameters described above, and a time series of prices that are sampled uniformly from $[20, 50]$ \$/MWh, which is shown in Fig. \ref{fig:data}.
The first $7296$ sets of data (corresponding to January to October) are used for training, and the last $1464$ sets of data for testing (corresponding to November and December). 
The neural networks are implemented using Tensorflow \cite{abadi2016tensorflow}, trained with the  Adam optimizer with a learning rate of $0.001$ and training step of $10000$.

\subsection{Performance Comparison}

The mean absolute percentage error (MAPE), and the standard deviation of absolute error (SDAPE) are used to evaluate the performance of the learned model.

\subsubsection{Linear Function}

We first test the performance of dynamical DR model under direct approach.
Table \ref{table:linear} shows the results when the linear function is used in the direct approach.
Note that when order $n=0$, i.e., no information on previous time interval is utilized, both the training error and testing error are high, with MAPEs being $19.67\%$ and $16.46\%$, respectively.
The errors drop significantly when $n > 0$, which verifies the claim that the price responsive characteristics of the EUCs depend on their states in previous time intervals, rather than being static over time.
Increasing the order significantly improves the model performance when $n < 3$, however, When $n\geq 3$, the improvement becomes negligible. Therefore, when the linear function is used, an appropriate order of the dynamical DR model would be $3$.
Intuitively, this means the response of the aggregate EUC demand to the price signal depends on the previous $3$ hours, rather than being static.

\begin{table}[!t]
	\caption{Performance of Dynamic DR Model Using Linear Function.}
	\label{table:linear}
	\centering
	\begin{tabular}{ccccccccc}
		\toprule
		& order $n$ & 0 & 1 & 2 & 3 & 4 & 5\\
		\midrule
		\multirow{2}*{\rotatebox{90}{Train}}
		& MAPE (\%) & 19.67 & 6.19 & 4.46 & 3.99 & 3.95 & 3.82\\
		& SDAPE (\%) & 15.22 & 4.97 & 3.89 & 3.61 & 3.56 & 3.46\\
		\midrule
		\multirow{2}*{\rotatebox{90}{Test}}
		& MAPE (\%) & 16.46 & 6.02 & 4.35 & 4.16 & 4.29 & 4.10\\
		& SDAPE (\%) & 13.91 & 4.80 & 3.55 & 3.38 & 3.43 & 3.31\\
		\bottomrule
	\end{tabular}
\end{table}

\subsubsection{FNN}

A FNN with $L=2$ hidden layers, each with $32$ neurons, is also used to represent the dynamical DR model.
Table \ref{table:FNN} shows the results when the FNN is used in the direct approach.
It is clear that the FNN outperforms the linear function when the order of the model is the same, which indicates that the linear function is underfitting the price and energy consumption data.
In addtion, when the FNN is used in the direct approach, the number of order of the dynamical DR model can be much lower than the case with the linear function.
For example, the testing MAPE of the FNN is $3.46\%$ when $n = 2$, much better than that of the linear function when $n = 5$, which is $4.10\%$.

\begin{table}[!t]
	\caption{Performance of Dynamic DR Model Using FNN.}
	\label{table:FNN}
	\centering
	\begin{tabular}{ccccccccc}
		\toprule
		& order $n$ & 0 & 1 & 2 & 3 & 4 & 5\\
		\midrule
		\multirow{2}*{\rotatebox{90}{Train}}
		& MAPE (\%) & 17.54 & 4.73 & 2.98 & 2.95 & 2.92 & 2.87\\
		& SDAPE (\%) & 12.45 & 3.64 & 2.55 & 2.46 & 2.45 & 2.38\\
		\midrule
		\multirow{2}*{\rotatebox{90}{Test}}
		& MAPE (\%) & 15.48 & 5.26 & 3.46 & 3.47 & 3.48 & 3.34\\
		& SDAPE (\%) & 11.74 & 3.77 & 2.74 & 2.64 & 2.64 & 2.63\\
		\bottomrule
	\end{tabular}
\end{table}

\subsubsection{RNN/LSTM}

In the indirect approach, a RNN/LSTM with $L=1$ hidden layer and $32$ neurons is adopted.
The results are presented in Table \ref{table:FNN}.
In the indirect approach, only the most recent price and energy consumption is sent into the RNN/LSTM for prediction; The testing MAPEs of the indirect approach, which are $3.25\%$ with RNN and $3.16\%$ with LSTM, are better than that of the direct approach with FNN of order $5$, which is $3.34\%$, while the LSTM performs better than the RNN by a small margin.

\begin{table}[!t]
	\caption{Performance of Dynamic DR Model Using RNN/LSTM.}
	\label{table:RNN}
	\centering
	\begin{tabular}{cccc}
		\toprule
		&  & RNN & LSTM\\
		\midrule
		\multirow{2}*{\rotatebox{90}{Train}}
		& MAPE (\%) & 3.02 & 2.83\\
		& SDAPE (\%) & 2.68 & 2.56\\
		\midrule
		\multirow{2}*{\rotatebox{90}{Test}}
		& MAPE (\%) & 3.25 & 3.16\\
		& SDAPE (\%) & 2.55 & 2.47\\
		\bottomrule
	\end{tabular}
\end{table}

\begin{figure}[!t]
	\centering
	\includegraphics[scale=0.6]{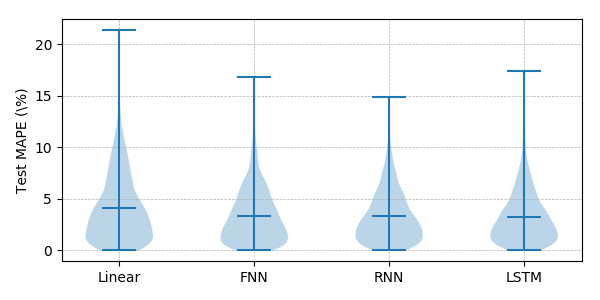}
	\caption{Testing MAPE under different approaches.}
	\label{fig:violin}
\end{figure}

The testing MAPE obtained using linear function with order $5$, FNN with order $5$, RNN, and LSTM are presented in the violin plot in Fig. \ref{fig:violin}.
To sum up, both the direct and indirect approach can be applied to learn a good dynamical DR model.
FNN, RNN, and LSTM outperforms the linear function, at the cost of higher model complexity.
In particular, when RNN or LSTM is used, no manual selection of states are required.

\subsection{Discussion}

When the neural networks are used to represent the dynamical DR model, we also investigated the impacts of hyperparameter such as the order the dynamical DR model, the number of layers, and the number of neurons in each layer.
Among all factors, the order of the dynamical DR model has the most significant impact on the performance, which has already been discussed in details in the above section.
The impacts of other parameters are relatively small.

Also, we would like to emphasize that the learned dynamical DR model can find its application in a variety of scenarios.
For example, it can be used to predict the energy consumption profile over several time intervals, given the prices over the same intervals.
It can also be used in an agent that simulates the collectively behavior of a set of EUCs without necessarily modeling all the underlying components; such an agent can be very helpful in developing certain algorithms.

\section{Concluding Remarks} \label{sec:con}

In this paper, we proposed a dynamical DR model that captures the temporal behavior of EUCs.
A key element in the proposed model is the choice of state, which can be determined using either a direct approach that selects the state manually from the available information, or an indirect approach that computes the states from input information.
The dynamical DR model can be represented by a linear function or FNN when the direct approach is used, and a RNN or LSTM network when the indirect approach is used.
Both these two approaches can achieve small MAPE when predicting the response of the aggregate energy consumption to the price.
Numerical simulation results validated that the dynamical DR models are indeed necessary to model the price response characteristics of the EUCs, which are inherently temporally correlated.

Future research will utilize the proposed dynamical DR model to learn a pricing policy of the LSE purely from historical data using agent-based algorithms.

\bibliographystyle{IEEEtran}
\bibliography{ref}

\end{document}